\documentclass[journal,comsoc]{IEEEtran}
\usepackage{amsmath,amsfonts}
\usepackage{algorithmic}
\usepackage{algorithm}
\usepackage{array}
\usepackage[caption=true,font=normalsize,labelfont=sf,textfont=sf]{subfig}
\usepackage{textcomp}
\usepackage{stfloats}
\usepackage{url}
\usepackage{verbatim}
\usepackage{graphicx}
\usepackage{cite}

\usepackage{xcolor}
\newcommand{\rebuttal}[1]{\textcolor{blue}{#1}}

\makeatletter
\def\endthebibliography{%
  \def\@noitemerr{\@latex@warning{Empty `thebibliography' environment}}%
  \endlist
}
\makeatother
\begin{document}

\title{Federated Learning for Internet of Things: Applications, Challenges, and Opportunities}

\author{Tuo Zhang, Lei Gao, Chaoyang He, Mi Zhang, Bhaskar Krishnamachari, and Salman Avestimehr}



\maketitle

\begin{abstract}
Billions of IoT devices will be deployed in the near future, taking advantage of faster Internet speed and the possibility of orders of magnitude more endpoints brought by 5G/6G. With the growth of IoT devices, vast quantities of data that may contain users' private information will be generated. The high communication and storage costs, mixed with privacy concerns, will increasingly challenge the traditional ecosystem of centralized over-the-cloud learning and processing for IoT platforms. Federated Learning (FL) has emerged as the most promising alternative approach to this problem. In FL, training data-driven machine learning models is an act of collaboration between multiple clients without requiring the data to be brought to a central point, hence alleviating communication and storage costs and providing a great degree of user-level privacy. However, there are still some challenges existing in the real FL system implementation on IoT networks. In this paper, we will discuss the opportunities and challenges of FL in IoT platforms, as well as how it can enable diverse IoT applications. In particular, we identify and discuss seven critical challenges of FL in IoT platforms and highlight some recent promising approaches towards addressing them.
\end{abstract}


\rebuttal{}

\section{Introduction}

The rapid advancement and expansion of the Internet of Things (IoT) result in exponential growth of data being generated at the network edge. 
Such advancement and expansion pose new challenges to the conventional cloud-based centralized approaches for data analysis from primarily two aspects. 
First, the centralized approaches no longer fit the 5G/6G era due to the extremely high communication and storage overhead (e.g., high-frequency data from high-volume time-series sensors such as video cameras or Lidar sensors) for pooling data from millions or billions of IoT devices. 
Second, the data being collected is increasingly viewed as threatening user privacy. With the cloud-based centralized approaches, user data could be shared between or even sold to various companies, violating privacy rights and negatively affecting data security, further driving public distrust with data-driven applications.
Therefore, a distributed privacy-preserving approach for data-driven learning and inference-based applications is needed for efficiency and to alleviate privacy concerns.

\begin{figure}[t]
\centering
\includegraphics[width=0.75\columnwidth]{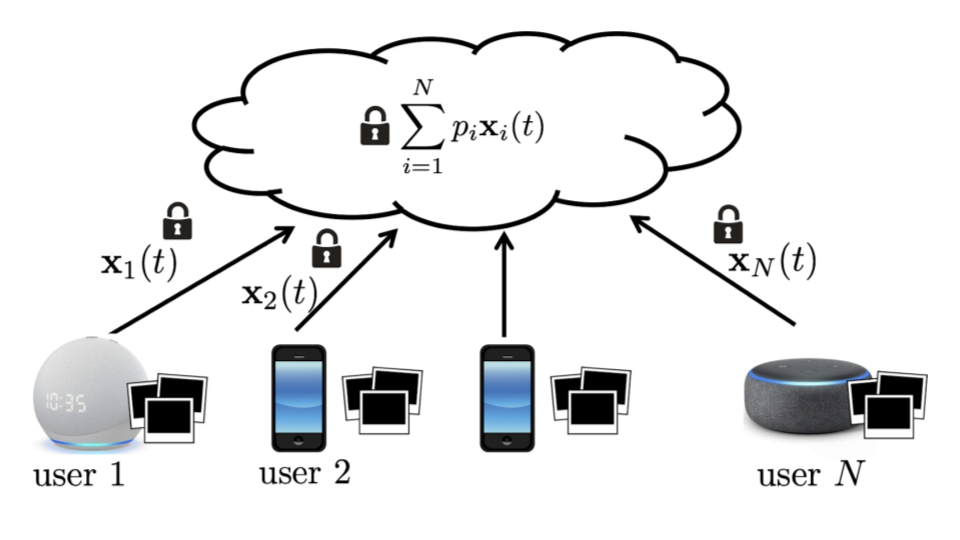}
\caption{Federated Learning for Internet of Things (IoT).}
\label{fediot-framework}
\end{figure}

In recent years, federated learning (FL) has emerged as a distributed privacy-preserving solution to addressing this pressing need.
The term federated learning was first introduced in 2016 by McMahan et al. \cite{federated-learning}. As shown in Figure~\ref{fediot-framework}, in FL, training of machine learning models for data-driven applications is an act of collaboration between distributed clients without centralizing the client data. 
The distributed and collaborative nature of FL is a natural fit to the network edge where each IoT device at the edge is an individual client. Moreover, since the raw data collected at each IoT device are not transmitted to others, FL provides an effective mechanism to protect user privacy particularly in the IoT domain where IoT sensors could directly capture data about users that contain privacy-sensitive personal information.

In this article, we briefly explain the advantages that FL brings to the IoT domain and discuss some of the most important IoT applications enabled by these advantages. We then focus on discussing some of the outstanding challenges across systems, networking, security, practical issues in real-world deployments, and development tools that act as the key barriers of enabling FL for the IoT domain and the opportunities in tackling these challenges.
To distinguish our work from existing efforts such as \cite{Kairouz2021AdvancesAO,Lim2020FederatedLI,Li2020FederatedLC,Imteaj2022ASO}, we focus on new challenges as well as articulating known challenges from new perspectives which have not been discussed before. 
We hope that this article could inspire new research that turns the envisioned Internet of Federated Things into reality.

\section{Why Federated Learning for IoT?}

\begin{figure}[t]
\centering
\includegraphics[width=0.57\columnwidth]{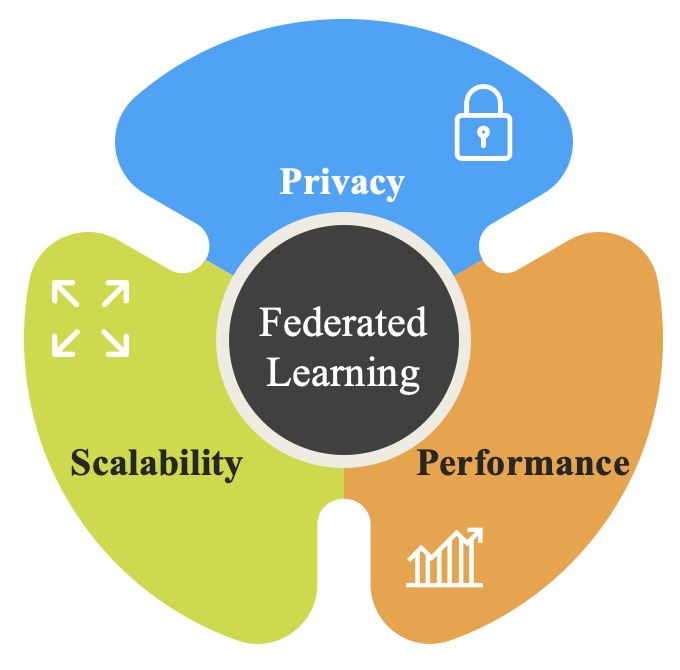}
\caption{Advantages of Federated Learning for IoT.}
\label{fed-advantages}
\end{figure}

The distributed, collaborative, and privacy-preserving characteristics of FL bring a number of key advantages for IoT applications (Figure~\ref{fed-advantages}) as follows:
\begin{itemize}
    \item \textbf{Preserving the Privacy of User Data}: In an ideal FL scenario, each IoT device in the system would learn nothing more than the information needed to play its role. The raw data never leaves the devices during the federated training process, and only the updates of the model are sent to the central server, which minimizes the risk of personal data leakage.
    \vspace{1mm}
    \item \textbf{Improving Model Performance}: Due to device constraints, a single IoT device may not have sufficient data to learn a high-quality model by itself. Under the FL framework, all the IoT devices can collaboratively train a high-quality model such that each participant could benefit from learning data collected by others beyond its own data but without probing others' private information. Moreover, as the FL could update the local model periodically, the edge device could always update its model in a time-varying manner. Thus, FL is an effective mechanism to enhance the model performance that each individual device cannot achieve by itself.
    \vspace{1mm}
    \item \textbf{Flexible Scalability}: The distributed nature of FL is able to leverage the constrained computation resources located at multiple IoT devices across different geographical locations in a parallel manner. As edge device hardware capability is increasing, the data size of each individual becomes huge, and centralizing all data to the server either wastes the computing resource at the edge or brings pressure for wireless communication network, which become an obstacle for the network scalability. By attracting more devices to join the framework, FL enhances the scalability of IoT networks without adding an extra burden on a centralized server due to its distributed learning nature. In addition, within the FL framework, there is no need for the expansive transmission of raw IoT-collected data, which also increases the scalability with regard to communication costs, especially for the low bandwidth IoT networks. 
    
\end{itemize}

\section{Applications}
\begin{figure}[t]
\centering
\includegraphics[width=0.95\columnwidth]{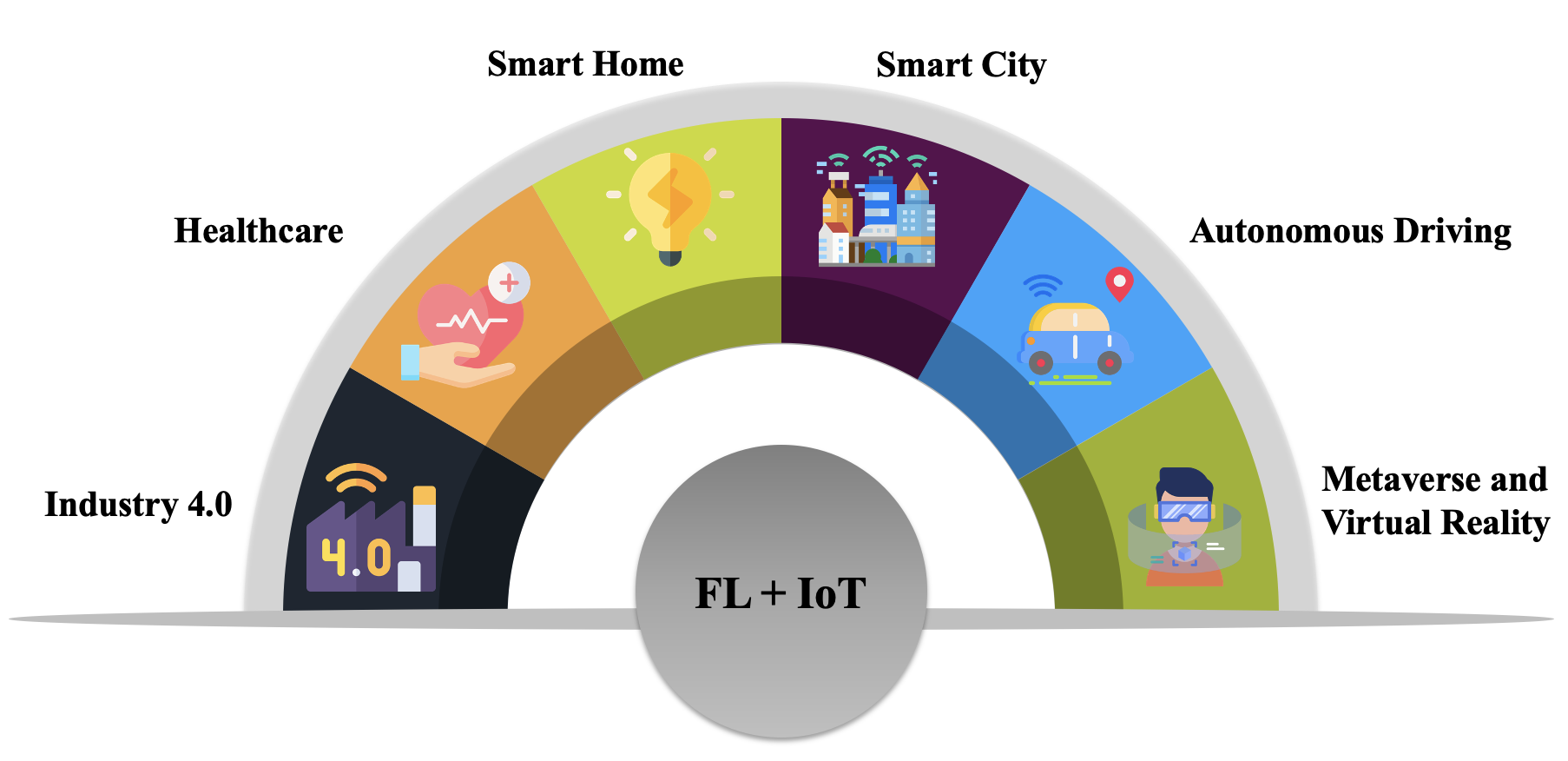}
\caption{Applications of Federated Learning for IoT.}
\label{fed-applications}
\end{figure}

Benefited from the advantages mentioned above, FL has enabled many important IoT applications. In this section, we briefly discuss some of the most important ones (Figure~\ref{fed-applications}).

\subsection{Industry 4.0}
The rapid development in the Industrial Internet of Things (IIoT) brings several advances in information technology applications for the manufacturing field. The concept of Industry 4.0, also known as the fourth industrial revolution, has been proposed based on the emergence of significance for the inter-connectivity of IIoT and the access to real-time data.
With unprecedented connectivity, Industry 4.0 will bring greater insight, control, and data visibility for the supply chain in many industries. Currently, some mature implementations of Industry 4.0 include the Optical Character Recognition (OCR) for the labels, smart and automatic Incoming Quality Control (IQC), and smart Process Quality Control (PQC). 
However, there are still some real-world problems challenging the deployment of Industry 4.0. First, the amount of data generated from a single factory may not be sufficient enough for training a reliable model comprehensively. Second, the data collected by the industrial IoT devices is highly related to the commercial value, which makes privacy-preserving important. For example, eavesdroppers may infer the capacity for manufacture from its electricity usage for industrial IoT users. The federated learning framework will become an inspired solution to address the above challenges.

\subsection{Healthcare}
As IoT devices become more pervasive in individuals' daily lives, the privacy of the collected data becomes a significant matter. An example to illustrate privacy concerns is IoT E-health. Nowadays, smart wearable devices are used to monitor the health status of patients, such as heartbeat, blood pressure, and glucose level. Compared to the other types of data, personal healthcare data is most sensitive to users' privacy and highly restricted by government laws and regulations for any kind of data sharing. Therefore, techniques such as FL are a requirement for investigators and researchers to develop state-of-the-art ML models over a fractured and highly regulated data landscape. The ability to train machine learning models at scale across multiple medical institutions without pooling data is a critical technology to solve the problem of patient privacy and data protection. Successful implementation of federated learning in healthcare could hold significant potential for enabling precision medicine at a large scale, helping match the right treatment to the right patient at the right time.

\subsection{Smart Home}
Smart home systems enabled by consumer IoT devices have achieved great popularity in the last few years as they improve the comfort and quality of life for the residents.
The wireless smart IoT home devices, such as smart bulbs, smart doorbells, and smart cameras, are capable of communicating with each other and controlled remotely by smartphones and microcontrollers. The implementations of Wake-Up-Word speech recognition and Automatic Speech Recognition (ASR) on IoT devices bring great convenience to everyday living, and people now tend to rely on smart IoT gateways with intelligent virtual assistants to control their home hands-free. 
FL has thus become a critical technology that is able to improve the on-device speaker verification while reducing the risk of raw data leakage.

\subsection{Smart City}
IoT-enabled smart cities are bringing significant advancements by making city operations efficient while improving the quality of life for citizens.
Various IoT devices enable city managers to control the physical objects in real-time and provide intelligent information to citizens in terms of the traffic system, transportation, public safety, healthcare, smart parking, smart agriculture, and so on. 
Due to the concerns of data privacy, smart infrastructures are moving to compute resources close to where data reside, which makes FL framework suitable for deployment. 
For example, the FL-based smart grid system enables collaborative learning of power consumption patterns without leaking individual power traces and contributes to the establishment of an interconnected and intelligent energy exchange network in the city.

\subsection{Autonomous Driving}
Along with the advancement of the vehicular IoT, autonomous driving technology is making its way into everyday cars. For a reliable self-driving system, it needs frequent real-time communication with a multi-access communication environment. Also, the spatial and temporal changes of the vehicular environment require an intelligent approach that can evolve with the change of environment.
For the traditional centralized-over-cloud method, the driving system needs to transmit a large amount of raw data to the server, which would cause potential privacy leakage. The communication overhead triggered by the large-size data transmission and limited network bandwidth may also lead the driving system to not being able to respond to the real-time spatial changes precisely. Adopting federated learning in vehicular edge computing for autonomous driving has thus become a promising direction to mitigate the above challenges. With FL, each vehicle only needs to transmit a limited size of data to the cloud and can adapt to real-time local changes more sensitively.

\subsection{Metaverse and Virtual Reality}
The metaverse is a hypothesized next generation of the internet, providing fully connected, immersive, and engaging online 3D virtual experiences through conventional personal computing, as well as virtual and augmented reality devices. In metaverse, users own their avatars and can interact with virtual objects and other participants. 
One of the fundamental building blocks of the metaverse is the digital twins duplicated in virtual environment that reflect the real-time physical world status. The connection between the virtual and physical world is tied by the data collected from IoT devices.
Federated learning is a promising solution to enable collaboration between edge and server for better global performance and also boost the security and privacy of the metaverse. For example, the eye tracking or motion tracking data collected by the wearables of millions of users can be trained in local devices and aggregated via an FL server. Hence, users can enjoy services in the metaverse without leaking their privacy.

\begin{figure}[t]
\centering
\includegraphics[width=0.7\columnwidth]{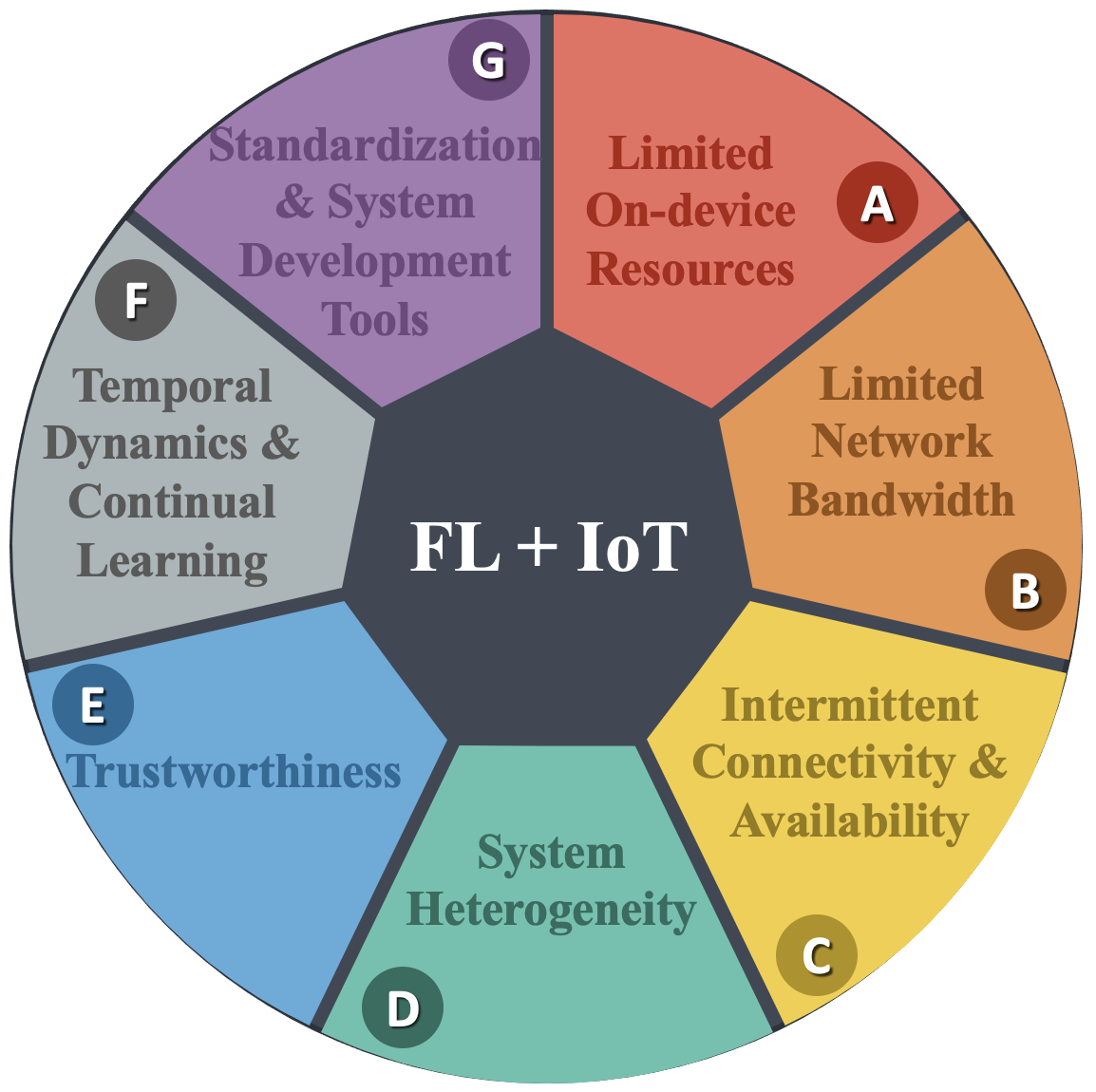}
\caption{Challenges of Federated Learning for IoT.}
\label{fed-challenges}
\end{figure}

\section{Challenges and Opportunities}
To realize the full potential of FL in the applications mentioned above, we have identified seven challenges that act as the key barriers of enabling FL on potentially billions of IoT devices. These challenges come from 1) the limited resources of the IoT devices, 2) limited network bandwidth available at the edge, 3) intermittent connectivity and availability commonly occurred in real-world settings, 4) the diversity of IoT devices across their available resources, 5) the temporal dynamics after deployments, 6) how to protect from the adversarial attacks and aggregate client information securely, and 7) the lack of standardization and system development tools in the community.
In the following, we describe these challenges followed by the opportunities that have high promise to address those challenges.

\subsection{Limited On-device Resources}
The deployment of FL on the network edge is severely impeded by the limited resources of the IoT devices. Existing machine learning models, especially deep neural networks, are known to be computation-intensive, which presents strict requirements on hardware and may result in low training efficiency on edge devices. Thus, developing customized and specialized hardware for machine learning applications on the edge is a promising direction to accelerate inference and training tasks while using much less energy compared with general-purpose processors.
Edge devices have limited resources not only in terms of computation but also in terms of memory for storage and data access. Recent neural network architectures require accessing a vast amount of memory locations for storing not only model weights and parameters but also the intermediate results produced by the computations. Therefore, a significant challenge for processing neural network models on a resource constraint device is reducing the memory accesses and keeping the data on-chip as to avoid costly reads and writes to the external memory modules.
Finally, in contrast to servers with CPUs and GPUs that can use a substantial amount of power, edge devices with embedded processors have a limited energy budget, which further imposes restrictions on the hardware performance. Despite the fact that current edge devices are increasingly powerful, training some deep learning models on-device is still time-consuming and inefficient.

To make models more applicable for the edge environment, researchers mainly focus on two research directions: design lightweight and hardware-friendly models/algorithms, and compress existing models to obtain thinner and smaller models, which are more computation and energy-efficient.
As an example, FedMask \cite{fedmask} is proposed as a joint computation and communication-efficient FL framework. By applying FedMask, each device can learn a heterogeneous and structured sparse binary mask; based on the mask, it is able to generate a sparse model with reduced computation cost, memory footprint, and energy consumption. However, this approach is hardware-agnostic; to further reduce the resource demands of federated training, we envision that the approach of hardware and algorithm co-design, which sparsifies the model by taking the IoT hardware architecture into consideration during federated training, is a promising future direction.

\subsection{Limited Network Bandwidth}
Communication bottleneck is considered one of the major challenges in an FL-based IoT environment.
Currently, most IoT devices communicate using wireless networks whose bandwidth is much smaller than wired network bandwidth in datacenters. For example, under a smart home scenario, the sum of the overall networking bandwidth is constant for the whole IoT system, no matter how many devices are connected. As more devices join the system, the communication problem is advent when clients possess different resources allocations. The limited network bandwidth not only makes the communication between clients and the server inefficient but also triggers the presence of straggler clients, which fail to share their local updates with the server during the communication round. They both serve as the bottlenecks for the performance of FL deployment in the large-scale IoT scenario.

To reduce the bandwidth demand during federated training, methods such as gradient compression have been heavily explored. However, these methods compromise the training quality to gain training efficiency.
Mercury \cite{Mercury} proposed a sampling-based framework that enables efficient on-device distributed training without compromising the training quality as a new inspiration for solving this challenge. 
In addition, Chen et al.’s work \cite{joint} formulates the bandwidth resource allocation and user selection problem during training federated learning models as an optimization problem whose goal is to minimize the training loss while meeting the delay and energy consumption requirements.
Liu et al. also proposed a client-edge-cloud hierarchical aggregation framework as a communication resource-efficient method to operate the federated learning in edge computing \cite{hagg}. Each client is able to offload its data samples and learning tasks from its device to the edge in proximity (e.g., edge gateway at home) for fast computation in the client-edge-cloud paradigm, which allows multiple edge servers to perform partial model aggregation. These works proposed promising and orthogonal techniques to reduce the bandwidth demand in the context of IoT. We envision that those techniques can be combined together in the scenario where IoT devices are confronted with extremely limited network bandwidth.


\subsection{Intermittent Connectivity and Availability}
Apart from the previous challenge of bandwidth limitations, the intermittent connectivity of the IoT devices signifies an unstable network connection that drops the device out of the system in the middle of the training round. Especially in large-scale IoT systems, the dropout problem followed by the intermittent connectivity and availability of various devices will become a serious obstacle for the FL framework to efficiently manage and schedule clients.
Currently, most of the FL studies are based on the synchronous update at the server, which implies that the server will not start the model aggregation until it receives the information sent from the slowest client. However, in real-world settings, due to the unbalanced communication abilities and training data distribution, the local training speed varies from device to device, and even some clients will be temporarily disconnected during the training phase, which makes the synchronous update nearly impossible. Also, not all of them will be simultaneously available for FL updating. In the asynchronous FL scenarios, a client could join the active learning group even in the middle of the training progress, which endangers the convergence of the federated training. 

To address this challenge, some researchers proposed an asynchronous aggregation scheme with the implementation of coding theory to resist the stragglers in the FL system. In \cite{Nguyen2021FederatedLW}, an asynchronous aggregation protocol known as FedBuff has been proposed to mitigate stragglers and enable secure aggregation jointly. Specifically, the individual updates are not incorporated by the server as soon they arrive. Instead, the server will keep receiving local updates in a secure buffer of size K, which is a tunable parameter, and then update the global model when the buffer is full. However, in real-world settings, IoT devices are by nature heterogeneous with diverse computing speeds. IoT devices with higher computing speed would be able to send in their local updates faster than IoT devices with slower computing speed, which inevitably leads to training bias. We envision that an asynchronous approach that can take the heterogeneity of IoT devices into account could be a better and promising solution.

\subsection{System Heterogeneity}
Within cross-device settings, clients under the FL framework have diverse system metrics in terms of both hardware and software. 
Various devices with different hardware architectures or even different device vendors are used to perform the learning tasks in different operating systems and different software APIs. Clients may choose different deep learning frameworks such as TensorFlow, PyTorch, and Caffe to train the local models, resulting in different model formats for aggregation. All the diversities have not only posed a significant challenge to system design but also exacerbated the asynchronous communication problem as mentioned above.
Moreover, in IoT settings, the data collected by different devices can be very different in terms of the feature and dimensions, and various types of devices can also have different temporal and spatial preferences for their data collection, which may create a discrepancy in the local data structure among all the participants under FL framework. For example, a surveillance camera will record videos in real-time (24x7 hours), while the data generated by a doorbell is intermittent. However, the central server could not examine the impact of the data heterogeneity until the training is done. 

An FL framework for IoT should enable graceful adaptation of the data and compute load across different devices based on their resource availability. To address this challenge, we envision that the training quality and speed will be improved if we can determine the heterogeneity and make adjustments accordingly before the training starts. Diao et al. \cite{Diao2021HeteroFLCA} proposed a heterogeneous FL framework that can produce a single global inference model from training heterogeneous local models on the clients. It is the first time to challenge the underlying assumption of existing work that local models have to share the same architecture as the global model, which inspires a solution to address the system heterogeneity among IoT devices.

\subsection{Temporal Dynamics and Continual Learning}
IoT sensing devices will, by their very nature, continuously collect new data, which will be used to update the model for lifelong or continual learning. With the objective of keeping providing services accommodating to newly collected data, the continued model update training poses a new challenge for resource-limited IoT devices. Specifically, as most of the IoT devices are memory-limited, their memory resources are not sufficient enough to handle both model inference and training. 
Furthermore, the lack of sufficient memory to keep past collected data may exacerbate catastrophic forgetting, which is one of the most critical problems in continual learning.

To address this challenge, we envision that the light-weighted machine learning engine is needed to reduce the memory consumption for on-device training. As an example, FedGKT \cite{He2020GroupKT} is a potential method to reduce the training memory footprint for efficient on-device learning. With FedGKT, IoT devices could transfer knowledge from many compact CNN models to a large CNN at a cloud server, which reformulates FL as a group knowledge transfer training model for the large-size model training on resource-constrained edge devices. To avoid catastrophic forgetting, we envision the use of clustering approaches that identify and store a few core data samples from each time interval. Moreover, leveraging IoT “Hubs” that can store non-sensitive/public datasets to inject memory in the training system is another promising solution. Furthermore, approaches that can detect temporal distribution shifts at each IoT node to determine when to update the model would also be needed.

\subsection{Trustworthiness}
%
In practical deployment, IoT devices are attractive targets for adversaries seeking to launch attacks such as phishing, identity theft, and distributed denial of service (DDoS). With the expansion of the IoT networks, the potential traffic volume of IoT-based DDoS attacks is reaching unprecedented levels, as witnessed during the Mirai botnet attack leveraging infected webcams and home routers. 
Attacks through the Internet have raised awareness of the need for IoT risk assessment and security, e.g., in fields such as healthcare. Even though these attacks could be easily defended by installing security patches, many IoT devices lack the requisite computation resources to do so. Moreover, within a cross-device system setting, it is difficult to identify whether the coming participant is malicious or not before it joins the system. Therefore, it is crucial for the IoT system to detect malicious or broken IoT devices that will ruin the model training with limited resources. 
To address this challenge, one of the promising directions is to implement a lightweight security protocol in the IoT system for the detection of broken and malicious devices. With the distributed nature, FL can offer an alternative approach for IoT cybersecurity by protecting the system from malicious attacks as close as possible to the IoT devices. DIoT \cite{diot} is the first system to employ an FL approach to anomaly-detection-based intrusion detection in gateways to IoT devices without centralizing the on-device data, where it demonstrates the efficacy of federated learning in detecting a wider range of attack types occurring at multiple devices.


Although FL shows its efficacy in cybersecurity for the IoT system, the privacy leakage from on-device sensitive data still matters as the participants share the model gradients or weight parameters with the server during the training process, which are derived from the participants' private training data as a statistical representation of the data it was trained on. The attacker could initiate the model inversion attack on the FL server first to achieve the individual model of each participant, and then recover the personal training data by inverting these personal models. One of the representative works for the attacking above is the inverting gradients attack \cite{inverting}, which proves that personal data reconstruction from gradient information is possible in federated learning setups. Therefore, a critical consideration in FL design is to ensure that the server as a blackbox for aggregation that does not learn the locally trained model of each user during model aggregation. Currently, the state-of-the-art secure aggregation protocols in FL essentially rely on two main principles: the pairwise random-seed agreement between users in order to generate masks that hide users’ models while having an additive structure that allows their cancellation when added at the server; and the secret sharing of the random-seeds so as to enable the reconstruction and cancellation of masks belonging to dropped users. The main drawback of such approaches is that the number of mask reconstructions at the server substantially grows as more users are dropped, causing a major computational bottleneck. Especially for the low-end IoT devices, the additional operator for the secure aggregation becomes an excessive burden to the limited on-device computational resources. To address this challenge, one of the promising directions is to implement lightweight and secure aggregation protocols that could provide the same level of privacy and dropout resiliency guarantees while substantially reducing the aggregation complexity, which meets the constraint in the IoT setting.

\subsection{Standardization and System Development Tools}
There are many concerns that the researchers need to take into account when designing a federated learning system on IoT networks. Issues such as different communication APIs, dataflow models, network configurations, and device properties have to be considered. As an emerging field, FL for IoT has not been standardized and appropriately implemented. Therefore, the research and development for standardization could help expedite the widespread deployment of FL systems on IoT networks and create an open environment for content sharing. Additionally, in light of the complexity involved in federated learning, researchers and industries need to further build upon existing FL developing and benchmarking tools such as TensorFlow Federated, PySyft, and FedML to accommodate the scenarios of IoT applications. From the application-level perspective,  user-friendly integrated simulation environments are needed to help design and evaluate the entire FL system on a large scale of IoT networks and its feasibility without implementing the model in real-world settings. From the system design perspective, ideally, we are looking for tools that can help developers accomplish system-level tasks such as load balancing, resource management, task scheduling, or data migrations easily.

One work of note along this direction is FedIoT\cite{fediot} which provides a mature systems-level framework that the developer can use to deploy their federated learning applications on CPU or GPU-enabled IoT devices, such as Raspberry Pi and NVIDIA Jetson Nano. To make federated learning more ubiquitous on IoT devices, we believe that researchers should pay attention to extending the current training frameworks to edge FL setting with awareness of the challenges mentioned above. It is worth mentioning that current edge computing solutions such as TensorFlow Lite, MNN, and TVM are focusing on improving the performance and efficiency of edge inference instead of training, much less taking FL setting into consideration, which is an under-explored area that would bring significant values to the federated learning and IoT communities.

\section{Concluding Remarks}
The distributed, collaborative, and privacy-preserving nature of federated learning makes it well suited for the IoT domain across a wide range of applications. 
In this article, we highlighted the key advantages and elaborated on some important applications of federated learning for IoT. We have also identified seven challenges that act as the key barriers of enabling FL for IoT followed by discussing opportunities to address these challenges. We hope this article acts as a catalyst to inspire new research at the intersection of federated learning and IoT.

\section{Acknowledgments}
This material is based upon work supported by Defense Advanced Research Projects Agency (DARPA) under Contract No. HR001120C0160, ARO award W911NF1810400, NSF grants CCF-1703575, CCF-1763673, CNS-2002874, ONR Award No. N00014-16-1-2189, a gift from Intel/Avast/Borsetta via the PrivateAI institute, a gift from Konica Minolta, and a gift from Cisco. The views, opinions, and/or findings expressed are those of the author(s) and should not be interpreted as representing the official views or policies of the Department of Defense or the U.S. Government. We would like to thank Eric van den Berg, Stuart Wagner, and Latha Kant from Peraton Labs for their insightful comments on an earlier draft of this work.

\bibliographystyle{IEEEtran}
\bibliography{aiot_survey}

\newpage

\section{Biography Section}

\vspace{-33pt}
\begin{IEEEbiography}[{\includegraphics[width=1in,height=1.25in,clip,keepaspectratio]{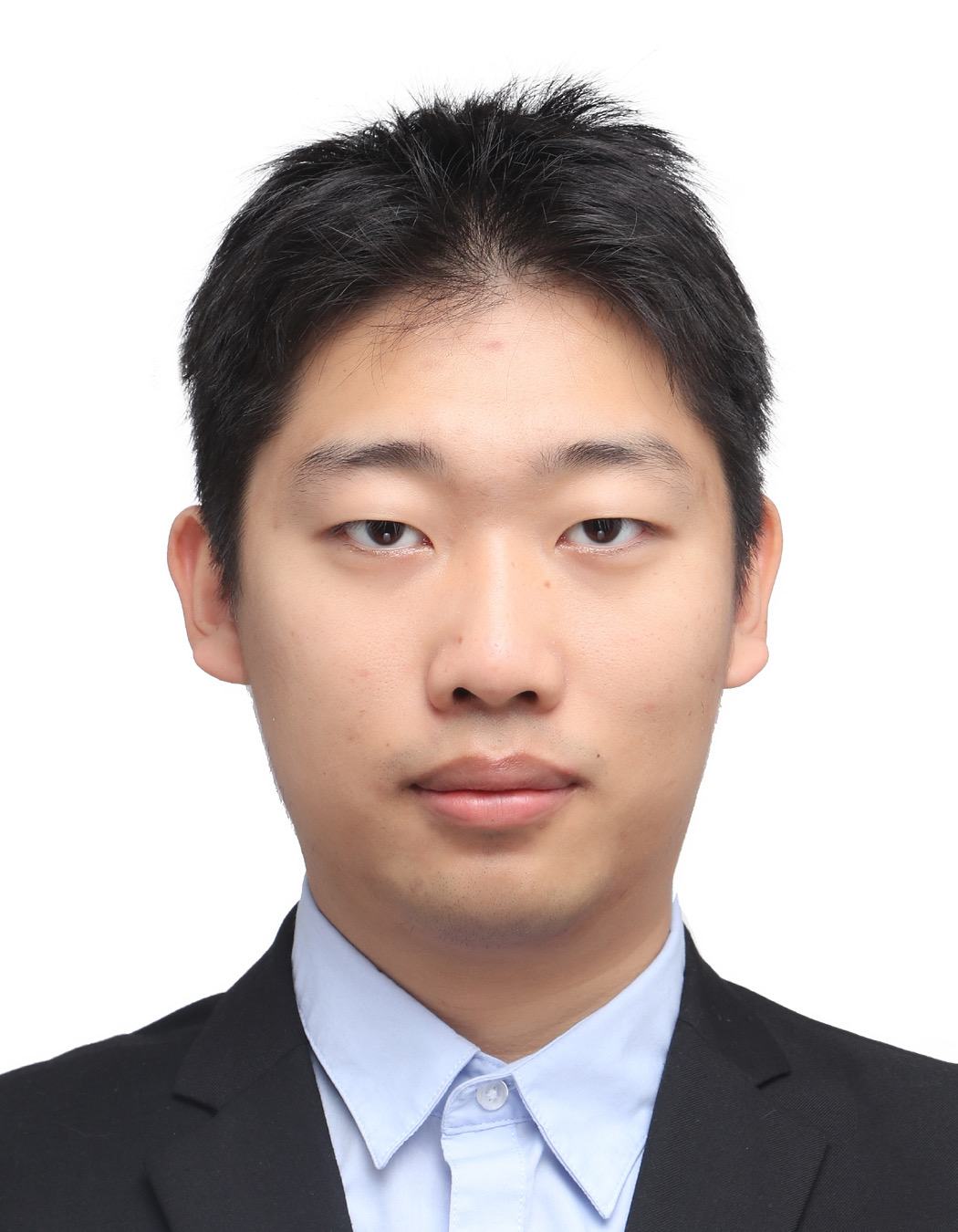}}]{Tuo Zhang}
Tuo Zhang (tuozhang@usc.edu) received his B.S. degree in electrical engineering from University of California, Santa Barbara in 2020. He is currently working toward a Ph.D. in Viterbi School of Engineering, University of Southern California. His research interest is in developing trustworthy machine learning algorithms and systems.
\end{IEEEbiography}

\begin{IEEEbiography}[{\includegraphics[width=1in,height=1.25in,clip,keepaspectratio]{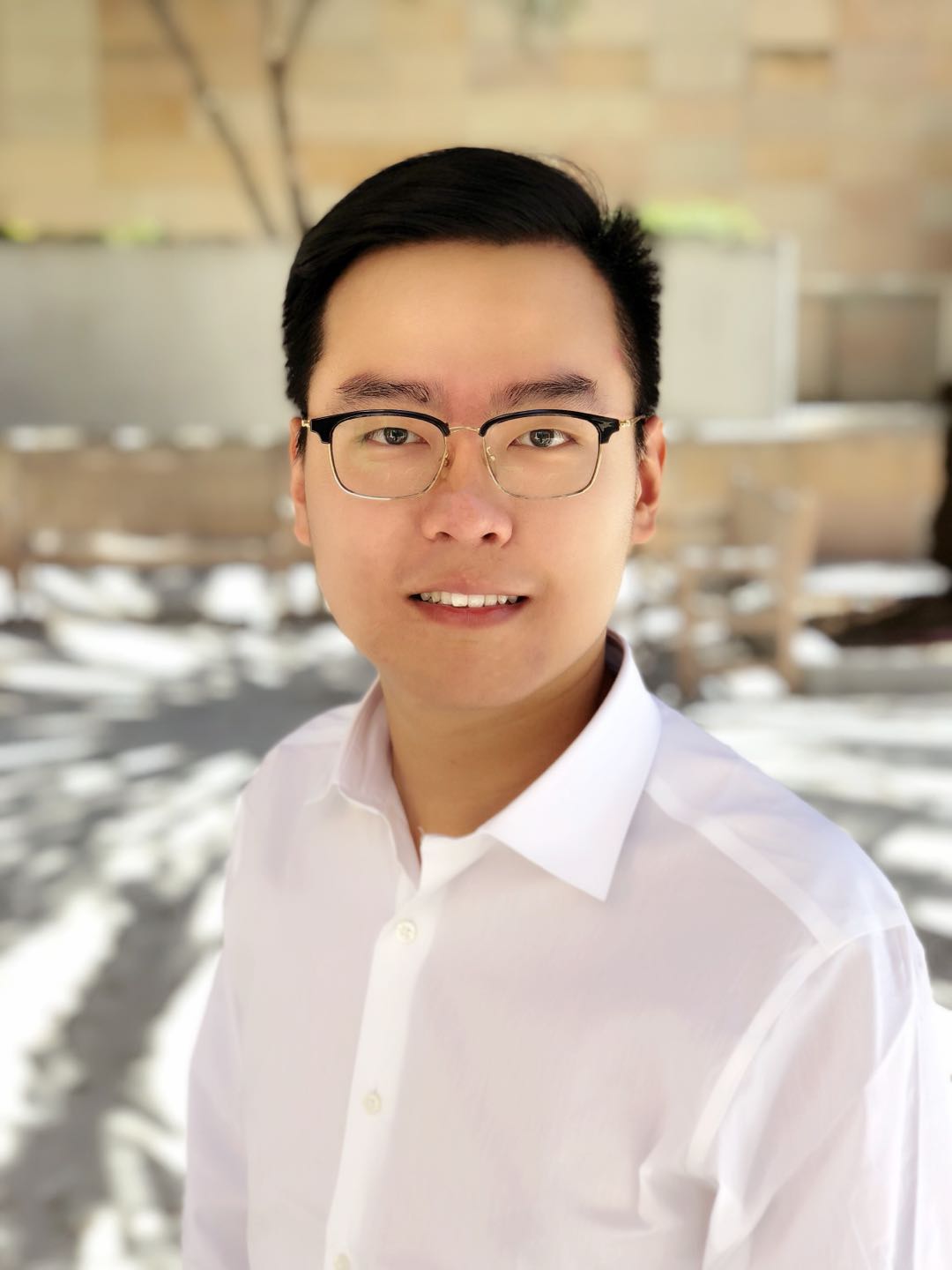}}]{Lei Gao}
Lei Gao (leig@usc.edu) received his B.S. degree in electrical engineering from University of California, Santa Barbara in 2019, and his M.S. in electrical engineering from University of Southern California in 2021. He is currently working as a student researcher at vITAL Lab in University of Southern California. His research interests are machine learning, Internet of Things and edge computing.
\end{IEEEbiography}

\begin{IEEEbiography}[{\includegraphics[width=1in,height=1.25in,clip,keepaspectratio]{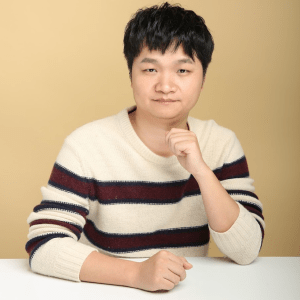}}]{Chaoyang He} Chaoyang He is a Ph.D. Candidate in the CS department at the University of Southern California. His research focuses on distributed/federated machine learning algorithms, systems, and applications. He is advised by Professor Salman Avestimehr (USC) and Mahdi Soltanolkotabi (USC). Previously, He was an R\&D Team Manager and Staff Software Engineer at Tencent (2014-2018), a Team Leader and Senior Software Engineer at Baidu (2012-2014), and a Software Engineer at Huawei (2011-2012). 

Chaoyang He has received a number of awards in academia and industry, including Best Paper Award at NeurIPS 2020 Federated Learning workshop, Amazon Machine Learning Fellowship (2021-2022), Qualcomm Innovation Fellowship (2021-2022), Tencent Outstanding Staff Award (2015-2016), WeChat Special Award for Innovation (2016), Baidu LBS Group Star Awards (2013), and Huawei Golden Network Award (2012). During his Ph.D. study, he has published papers at ICML, NeurIPS, CVPR, ICLR, MLSys, among others. His homepage: \url{https://chaoyanghe.com}.
\end{IEEEbiography}

\begin{IEEEbiography}[{\includegraphics[width=1in,height=1.25in,clip,keepaspectratio]{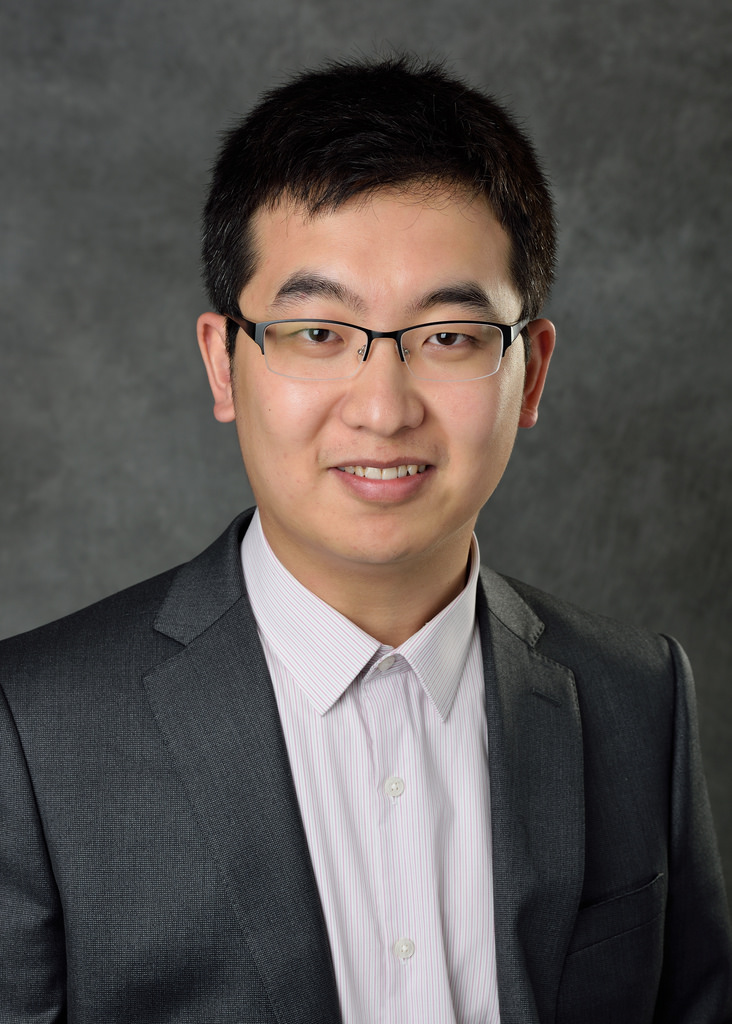}}]{Mi Zhang}
Mi Zhang is an Associate Professor and the Director of the Machine Learning Systems Lab at Michigan State University. He received his Ph.D. from University of Southern California and B.S. from Peking University. Before joining MSU, he was a postdoctoral scholar at Cornell University. His research lies at the intersection of mobile/edge/IoT systems and machine learning, spanning areas including On-Device AI, Automated Machine Learning (AutoML), Federated Learning, Systems for Machine Learning, Machine Learning for Systems, and AI for Health and Social Good.

Dr. Zhang has received a number of awards for his research. He is the 4th Place Winner of the 2019 Google MicroNet Challenge, the Third Place Winner of the 2017 NSF Hearables Challenge, and the champion of the 2016 NIH Pill Image Recognition Challenge. He is the recipient of seven best paper awards and nominations. He is also the recipient of the National  Science  Foundation CRII Award, Facebook Faculty Research Award, Amazon Machine Learning Research Award, and MSU Innovation of the Year Award.
\end{IEEEbiography}

\begin{IEEEbiography}[{\includegraphics[width=1in,height=1.25in,clip,keepaspectratio]{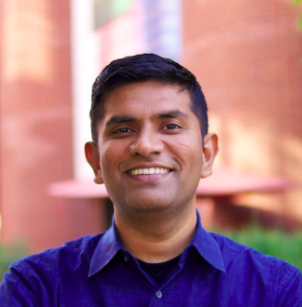}}]{Bhaskar Krishnamachari}
Bhaskar Krishnamachari is Professor of Electrical and Computer Engineering at the USC Viterbi School of Engineering. He is the founding director of the USC Viterbi Center for Cyber-Physical Systems and the Internet of Things. He received his M.S. and Ph.D. in Electrical Engineering from Cornell University in 1999 and 2002 respectively, and his B.E. in Electrical Engineering from The Cooper Union for the Advancement of Science and Art in 1998. His research interests pertain to the design and analysis of algorithms, protocols and applications for the internet of things, distributed computing, blockchain technologies, and networked robotics. He is the recipient of an NSF CAREER Award, the ASEE Terman Award, IEEE-HKN Outstanding Young Electrical and Computer Engineer Award, and several conference best paper awards including at ACM Mobicom and IEEE/ACM IPSN. 
\end{IEEEbiography}

\begin{IEEEbiography}[{\includegraphics[width=1in,height=1.25in,clip,keepaspectratio]{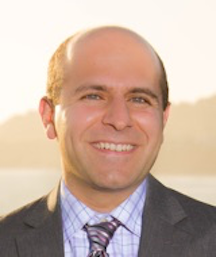}}]{A. Salman Avestimehr} A. Salman Avestimehr is a Dean's Professor, the inaugural director of the USC-Amazon Center on Secure and Trusted Machine Learning (Trusted AI), and the director of the Information Theory and Machine Learning (vITAL) research lab at the Electrical and Computer Engineering Department of University of Southern California. He is also an Amazon Scholar at Alexa AI. He received his Ph.D. in 2008 and M.S. degree in 2005 in Electrical Engineering and Computer Science, both from the University of California, Berkeley. Prior to that, he obtained his B.S. in Electrical Engineering from Sharif University of Technology in 2003. His research interests include information theory, large-scale distributed computing and machine learning, secure and private computing/learning, and federated learning.

Dr. Avestimehr has received a number of awards for his research, including the James L. Massey Research \& Teaching Award from IEEE Information Theory Society, an Information Theory Society and Communication Society Joint Paper Award, a Presidential Early Career Award for Scientists and Engineers (PECASE) from the White House (President Obama), a Young Investigator Program (YIP) award from the U. S. Air Force Office of Scientific Research, a National Science Foundation CAREER award, the David J. Sakrison Memorial Prize, and several Best Paper Awards at Conferences. He has been an Associate Editor for IEEE Transactions on Information Theory and a general Co-Chair of the 2020 International Symposium on Information Theory (ISIT). He is a fellow of IEEE.  \url{https://www.avestimehr.com}.
\end{IEEEbiography}

\vspace{11pt}

\vfill

\end{document}